 \documentclass[smallextended]{svjour3}       
\smartqed  
\usepackage{graphicx}
\usepackage{amsmath}
\usepackage{mathtools}
\usepackage{algorithm}
\usepackage{algorithmicx}
\usepackage{algorithm,algcompatible,amssymb,amsmath}

\usepackage{siunitx,etoolbox}

\usepackage{hyperref}

\newlength{\commentindent}
\renewcommand{\COMMENT}[2][.5\linewidth]{%
  \leavevmode\hfill\makebox[#1][l]{//~#2}}
\algnewcommand\algorithmicto{\textbf{to}}
\algnewcommand\RETURN{\State \textbf{return} }

\usepackage{natbib}
\begin{document}

\title{Relational Weight Priors in Neural Networks for Abstract Pattern Learning and  
Language Modelling
}


\titlerunning{Weight Priors in Neural Networks for Abstract Patterns and Language Models}        

\author{Radha Kopparti        \and
        Tillman Weyde 
}


\institute{Radha Kopparti \at
 \email{radha.kopparti@city.ac.uk}      
           \and
          Tillman Weyde \at
            \email{t.e.weyde@city.ac.uk}
}

\date{Received: May 15, 2020}

\maketitle

\begin{abstract}
Deep neural networks have become the dominant approach in natural language processing (NLP). 
However, in recent years, it has become apparent that there are shortcomings in
systematicity that limit the performance and data efficiency of deep learning in NLP. 
These shortcomings can be clearly shown in lower-level artificial tasks, mostly on synthetic data. 
Abstract patterns are the best known examples of a hard problem for neural networks in terms of generalisation to unseen data.  
They are defined by relations between items, such as equality, rather than their values.  
It has been argued that these low-level problems demonstrate the inability of neural networks to learn systematically. 

In this study, 
we propose Embedded Relation Based Patterns (ERBP) as a novel way to create a relational inductive bias that encourages learning equality and distance-based relations for abstract patterns. 
ERBP is based on Relation Based Patterns (RBP), but modelled as a Bayesian prior on network weights and implemented as a regularisation term in otherwise standard network learning.  
ERBP is is  easy to integrate into standard neural networks and does not affect their learning capacity.
In our experiments, ERBP priors lead to almost perfect generalisation when learning abstract patterns from synthetic noise-free sequences. 
ERBP also improves natural language models on the word and character level and pitch  prediction in melodies with RNN, GRU and LSTM networks. 
We also find improvements in in the more complex tasks of learning of graph edit distance and compositional sentence entailment. 
ERBP consistently improves over RBP and over standard networks, showing that it enables abstract pattern learning which contributes to performance in natural language tasks. 

\end{abstract}

\section*{Declaration}

\begin{itemize}
    \item \textbf{Funding} : Radha Kopparti is supported by a PhD studentship from City, University of London
    \item \textbf{Conflicts of interest/Competing interests} : No conflicts of interest
    \item \textbf{Availability of data and material} : All the datasets used in the experiments are available online
    \item \textbf{Code availability} : The code for the experiments will be made available online 
    \item \textbf{Authors' contributions} : Included in Introduction
\end{itemize}

\section{Introduction}
\label{sec:intro}
Humans are very good at learning abstract patterns, i.e. patterns based on relations rather than values, 
from sensory input and systematically applying them to new stimuli, even after brief exposure.
On the other hand, neural networks fail to detect such patterns in unseen data  \citep{GaryMarcus1999}.
The lack of systematicity in neural network learning has already been discussed over 30 years ago \citep{fodor1988connectionism}, but that discussion has still not led to a generally accepted consensus.
Nevertheless, deep learning has had impressive successes based on computational advances and greatly increased amounts of data. 
Yet, it has become evident that there are still relevant limitations to systematicity and compositionality of learning and generalisation in current neural network architectures \citep[e.g.,][]{lake2018generalization, GaryMarcus2018,hupkes2019compositionality,DBLP:journals/corr/abs-1904-00157,keresztury2020compositional}. 
Neural networks are effective at mapping numerical patterns similar to the training data to corresponding outputs, but often do not extrapolate this mapping successfully on unseen data \citep{liska-et-al-2018-memorize}.
Specifically, standard neural networks do not seem to learn equality relations \citep{mitchell2018extrapolation,weyde_kopparti_2018}, 
which define simple abstract patterns, such as those 
used in the well-known study by \cite{GaryMarcus1999}.

In this study, we use Embedded RBP (ERBP) \citep{weight2020arx}, an extension of Relation Based Patterns \citep{weyde_kopparti_2019}, for effective generalisation of equality and similarity relationships. 
ERBP is modeled as a prior over the network weights in standard feed-forward and recurrent network architectures. 
Thus, ERBP enables the use of existing standard architectures with just one additional term in the loss function, as opposed to other architectures addressing systematicity,  and it uses no fixed weights, in contrast to RBP. 
Our experimental results confirm that ERBP consistently leads to almost perfect generalisation on abstract pattern learning. 

\cite{GaryMarcus1999} and \cite{GaryMarcus2018} see the failure to learn abstract patterns as evidence for general limitations of neural network learning.
To our knowledge, this link has not been shown experimentally. 
Therefore we evaluate our solution  on language models for words and characters on real-life data. 
We find consistent improvements in all experiments, providing evidence for a link between the capacity for low-level abstraction and performance in natural language modelling. 

The main contributions of this study are: 
\begin{itemize}
    \item The application of the novel ERBP method for  weight priors that enable abstract pattern learning 
    \item Extensive experimental evaluation on abstract pattern learning with perfect accuracy on synthetic test data
    \item Multiple experiments on natural language modelling and graph neural networks, showing consistent improvements through ERBP
\end{itemize}

The remainder of the paper is organised as follows: 
Section \ref{sec:relatedwork} discusses the related work.
In Section \ref{sec:inductivebias}, learning of abstract patterns with ERBP is explained. 
Sections \ref{sec:abstract_pattern}, \ref{sec:realworld}, and \ref{exp:complex} describe the experimental setup and results on synthetic, real-world, and complex tasks, followed by the Discussion and Conclusions in Sections \ref{discussion} and \ref{conclusions}, respectively. 

\section{Related Work}
\label{sec:relatedwork}
There has recently been an increased interest in inductive biases to improve generalisation in machine learning and specifically neural networks with potential benefits in various applications like relational reasoning \citep{battaglia2018relational}, spatial reasoning \citep{hamrick2018relational}, learning from few examples \citep{snell2017prototypical,DBLP:journals/corr/abs-1904-05046}, cognitive modelling \citep{cog1}, natural language processing \citep{mitchell2018extrapolation}, machine translation \citep{sutskever2014sequence,lake2018generalization}, and numeric reasoning \citep{trask2018neural}. 

Earlier work on systematicity of human language learning and connectionism started with \citep{fodor1988connectionism}. 
There have been wider debates on systematicity of neural network learning with claims and counter-claims on their abilities \citep{fodor1990connectionism,niklasson1994systematicity,christiansen1994generalization,frank2014getting}.
In the context of abstract pattern learning, there was series of studies triggered by \cite{GaryMarcus1999} with different extended network architectures and evaluation methods \citep{Elman1999,Altmann2,vilcu2005two,shultz2006neural,Alhama,Alhama2019}, but no general consensus has emerged. 

Traditional rule based approaches 
do not suffer from a lack of systematicity, but they lack the flexibility of neural networks. 
Some approaches aim to improve rule based learning with more flexibility, e.g. through probabilistic logic \citep{deraedt,8618135}. 
Other studies aim at combining symbolic rule based knowledge with neural networks \citep{city11838,DBLP:journals/corr/abs-1711-03902,Raedt2019NeuroSymbolicN,doi:10.1098/rstb.2019.0309}. 
However, this approach has not yet been adopted in the mainstream of machine learning research and applications. 

There has been work on different ways of modelling 
inductive biases in neural networks in recent years, like using matrix factorisation \citep{DBLP:journals/corr/NeyshaburTS14} and convolutional arithmetic circuits   \citep{DBLP:journals/corr/CohenS16a} for computer vision tasks, relational inductive bias models and Bayesian inference in the context of  deep reinforcement learning \citep{zambaldi2018deep,Gershman2015NoveltyAI} and inductive bias for integrating tree structures in LSTMs by ordering neurons \citep{shen2018ordered}. 

It has been identified for a long time that weight initialisation can improve the speed and quality of neural network learning \citep{Sutskever2013OnTI,DBLP:journals/corr/LeJH15,nye2018efficient}.
Various regularisation methods have also been proposed to improve generalisation  \citep{DBLP:journals/corr/Sussillo14,DBLP:journals/corr/ZarembaSV14,Pachitariu2013RegularizationAN}.  
\cite{mikolov-et-al-2015-learning} use two weight matrices with different learning dynamics to encourage short and long term memory. 
Recently, spatial weight priors have been proposed for Bayesian convolutional neural networks by \cite{atanov2018the}.
There is work by \cite{demeester2016lifted} for injecting inductive bias in the form of implication first order rules through an additional regularisation term for learning relations between entities in WordNet. 

Recent studies have confirmed that state of the art neural networks lack systematic generalisation, specifically recurrent neural networks for a sequence to sequence learning task \citep{lake2018generalization,DBLP:journals/corr/abs-1802-06467,DBLP:journals/corr/abs-1809-04640}
and have stressed the lack of compositionality and generalisation in neural networks \citep{DBLP:journals/corr/abs-1906-05381,DBLP:journals/corr/abs-1904-00157,nye2020learning,keresztury2020compositional,doi:10.1098/rstb.2019.0309,DBLP:journals/corr/abs-1903-12354,hupkes2019compositionality,Andreas2019MeasuringCI}.

Other works on incorporating symbolic prior knowledge into neural networks include that by \cite{xu2018semantic}, where a loss function based on constraints on the output has been developed as a regularisation term and evaluated on structured prediction and multi-label classification tasks.
\cite{DBLP:journals/corr/MinerviniDRR17, inbook} proposed a method for regularising multi-layer graph based neural networks using logic based prior knowledge on link prediction tasks.
There are other works like Neural Theorem Provers (NTPs) \citep{DBLP:journals/corr/RocktaschelR17} which are proposed to solve large scale knowledge reasoning tasks \citep{minervini2019differentiable} and systematic generalisation \citep{minervini2020learning}.
In the context of relational reasoning and natural language understanding, works like \cite{sinha2019clutrr, sinha2020evaluating} evaluate systematic out-of-order logical generalisation in graph neural networks.

\cite{mitchell2018extrapolation} studied among other problems the learning equality of numbers in binary representation, which does not generalise from even to odd numbers as pointed out already by \cite{GaryMarcus2001}. 
They investigate several potential solutions, with a preference for a convolutional approach.  \cite{trask2018neural} studied the numerical extrapolation behaviour using logic gates based on arithmetic logic units for fundamental numerical operations.

In our previous work \citep{weyde_kopparti_2018,weyde_kopparti_2019,kopparti-weyde-2019}, we have shown that equality relations and abstract patterns based on such relations are not generalised by standard feed-forward and recurrent neural network architectures. 
This failure is independent of various parameters like the vector dimension, amount of training data, train/test split, type of activation function, data representation, vector coverage, and aspects of the network architecture. 
The proposed RBP model, 
creates a bias that leads to almost perfect  generalisation on a number of tasks that relate to equality detection and learning abstract patterns.
\cite{tanneberg2020learning} use the differential rectifier units proposed as part of RBP \citep{weyde_kopparti_2018}
to develop an architecture for learning efficient strategies that solve abstract algorithmic problems. 
In this work, we extend the RBP framework in the form of a Bayesian prior on the network weights to enable abstract relation learning.

\section{An Inductive Bias for Learning Abstract Patterns}
\label{sec:inductivebias}
To the best of our knowledge, a solution based on weight priors has not been proposed for the abstract pattern learning task. 
We therefore re-model the RBP approach by proposing a weight prior which makes it a more effective inductive bias and simpler to integrate into standard neural networks. 

\subsection{Motivation}
Neural networks are effective in recognising and mapping numerical patterns seen in the training data to corresponding outputs, but very often do not extrapolate this mapping outside
the range of the training data \citep{liska-et-al-2018-memorize}.
Specifically, standard neural networks do not seem to learn equality based abstract relations \citep{mitchell2018extrapolation}, 
which are fundamental for many higher level tasks. 
Learning equality or identity based relations is not only a very useful numerical relationship but also an important characteristic that defines abstract relations in data
e.g., the grammar-like rules ABA or ABB as 
used in the well-known study by \cite{GaryMarcus1999}.

It is not clearly understood why the standard neural network models in their current form are not able to learn abstract grammar-like rules based on equality.
To this end, we have proposed solutions based on \textit{Relation Based Pattern}s (RBP) \citep{weyde_kopparti_2018, weyde_kopparti_2019}, 
and \textit{Embedded Relation Based Pattern}s (ERBP) in \cite{weight2020arx} as re-modelling RBP as a weight prior with a regularisation term in otherwise standard neural network learning.

Here, we take this approach forward, and apply ERBP firstly to a wide range of synthetic tasks of learning abstract, mixed and concrete patterns, thereby modelling lower-level abstract patterns in data. 
We then aim to evaluate whether modelling the lower-level abstractions in data can improve higher level tasks like language modelling. 
To this end, we experiment with ERBP on real-word tasks of character and word prediction, music pitch prediction, graph distance modelling, and sentence entailment.
The aim behind this work 
is not to improve the state-of-the-art performance on these tasks but 
rather to focus on highlighting the problem of lower-level abstract pattern learning with neural networks and show that creating an inductive bias as a weight prior leads to improvements in several tasks. 

\subsection{Relation Based Patterns (RBP)}
The RBP model introduced in \cite{weyde_kopparti_2018,weyde_kopparti_2019} is based on the comparison of input neurons that 
are in some relation to each other. 
For the comparison,  Differentiator-Rectifier (DR) units are used that calculate the absolute difference of two inputs: $f(x,y) = |x-y|$. 
DR units realise this  
with fixed weights from the two neurons that are to be compared, with values $+1$ and $-1$, and the absolute value as activation function. 

For abstract pattern learning, there are multiple vector comparisons that correspond to the different possible relations within the patterns, e.g. equality of pairs of vectors representing tokens in positions (1,2), (1,3), (2,3), to recognise abstract patterns within three elements.
For each dimension of the input vectors and each comparison there is a $DR_n$ unit.
$DR_n$ units are applied to every pair of corresponding input neurons representing the same dimension within a token representation.
There are also $DRp$ units, which we do not model in ERBP, that can be used to aggregate the activations for all dimensions in a vector comparison by summing the activations of the $DR_n$ values. \cite{weyde_kopparti_2019} has a detailed description of $DR_{n}$ and $DR_p$ units.


\begin{figure}[tb!] 
\centering{\includegraphics[width=5cm]{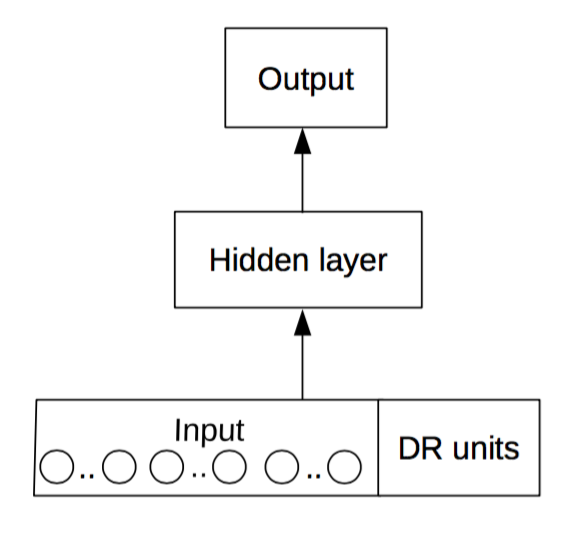}}
\caption{Structure of RBP Early Fusion. 
The DR units are concatenated with the input neurons. 
The fixed connections from the Inputs to the DR units are not shown for simplicity.}
\label{fig:rbp1} 
\end{figure}
\begin{figure}[tb!] 
\centering{\includegraphics[width=5cm]{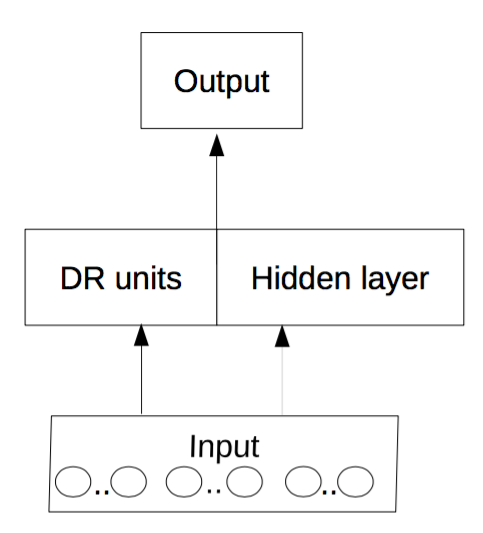}}
\caption{Structure of the RBP Mid Fusion. The DR units are concatenated with the first hidden layer. 
The weights of incoming connections of the DR units are fixed in RBP (see the main text for details). 
In ERBP, we use the same structure but with trainable weights. 
Then the only difference to a standard Feed Forward Neural Network is the prior on the weights between input and hidden units as described in the text. 
This structure can be applied to Recurrent Neural Networks, as long as we have $n$ items represented in the input.}
\label{fig:rbp2} 
\end{figure}

There are different ways of integrating DR units into neural networks in RBP: \textit{Early},  \textit{Mid} and \emph{Late Fusion}. 

In \emph{Early Fusion}, DR units are concatenated to input units as shown in Figure \ref{fig:rbp1}. 
In \emph{Mid Fusion}, they are concatenated to the hidden layer as shown in Figure \ref{fig:rbp2}. 
In both cases, the existing input and hidden units are unchanged. 
The \emph{Late Fusion} design 
predicts abstract relations and is more complex. 
Late Fusion are not used in this study, but a detailed description can be found in \cite{weyde_kopparti_2019}.



\subsection{Embedded Relation Based Patterns}

RBP adds units with hard-wired connections and non-standard activation function, which limits the flexibility of that part of the network and makes practical use more complex. 
By contrast, ERBP introduces a modified Mid Fusion RBP structure and remodels it as a Bayesian prior on a standard weight matrix, thus avoiding hard-coded weights and non-standard network structures and activation functions. 
The approach of using a prior has been introduced in our previous work \cite{weight2020arx} for equality learning and here we are extending it to abstract patterns and language modelling.

We model each DR unit of RBP with two standard neurons using ReLU activation and model the fixed weights with a default weight matrix $D$, that is applied at initialisation and through a loss function to the weights between input neurons and a hidden layer. 
ERBP therefore has two nodes of comparison per pair of input neurons, unlike RBP which has one, and we need therefore a higher minimal number of hidden units in the first layer.


The matrix is initialised between the input and the first hidden layer as follows.
The incoming connections from two corresponding inputs to a neuron in the hidden layer to be compared have values of $+1$ and $-1$. 
For the same pair we use another hidden neuron with inverted signs of the weights, as in rows 1 and 2. 
All other incoming weights to the hidden layer are set to 0, including the bias. 
This ensures that in all cases where corresponding inputs are not equal, there will be a positive activation in one of the hidden neurons. 


More formally, we can define the matrix $D$ as follows. 
Consider an architecture with input consisting of $n$ tokens represented as vectors of dimensionality $m$, concatenated to a $n \cdot m $ dimensional input vector. 
Then each token vector $i$ starts at position $i \cdot m$ (assuming $i$ starts at $0$). 
In the first hidden layer, we represent each of the $n(n-1)/2$ pairs of input tokens with $2m$ neurons, so that the representation of pair $j$ starts at position $p_j = j \cdot 2m$ (assuming 0-indexing again). 
The hidden layer has to have at least $n(n-1)/2 \cdot 2m = n(n-1)m$ neurons. 
We can then describe the creation of the matrix D with algorithm \ref{eq}.

\begin{algorithm}
 \caption{ERBP creation of weight matrix $D$}
 \label{eq}
 \begin{algorithmic}
    \STATE $D_{q,r} \leftarrow 0 \, \forall q,r$ \COMMENT{initialise the matrix to 0} 
    \STATE $p \leftarrow n(n-1)/2 $ \COMMENT{number of token pairs}
    \STATE $k \leftarrow 0$ \COMMENT{will be used as row index}
    \FOR{ $i_1$ from $0$ to $p - 1$} 
    \COMMENT{first token in pair}
    \FOR{ $i_2$ from $i_1 + 1$ to $p$} 
    \COMMENT{second token in pair}
    \FOR{ $j$ from 0 to $m$} \COMMENT{dimensions in token vectors}
         \STATE $D_{k, i_1 \cdot m + j} = +1$ \COMMENT{first comparison neuron}
         \STATE $D_{k, i_2 \cdot m + j} = -1$
         \STATE $k \leftarrow k+1 $
         \STATE $D_{k, i_1 \cdot m + j} = -1$ \COMMENT{second comparison neuron} 
         \STATE $D_{k, i_2 \cdot m + j} = +1$
         \STATE $k \leftarrow k+1 $
    \ENDFOR
   \ENDFOR
   \ENDFOR
 \end{algorithmic}
\end{algorithm}

An example of this default matrix $D$ for three tokens of length $m=2$ is shown in  Figure~\ref{fig:rbp4_2}. 
\begin{figure}[h]
\centering
{
$\displaystyle
D = {\text{\boldmath$
\begin{pmatrix} 
+1 & 0 &-1 & 0 & 0 & 0 \\
-1 & 0 &+1 & 0 & 0 & 0 \\
 0 &+1 & 0 &-1 & 0 & 0 \\  
 0 &-1 & 0 &+1 & 0 & 0 \\  
+1 & 0 & 0 & 0 &-1 & 0 \\
-1 & 0 & 0 & 0 &+1 & 0 \\
 0 &+1 & 0 & 0 & 0 &-1 \\  
 0 &-1 & 0 & 0 & 0 &+1 \\  
 0 & 0 &+1 & 0 &-1 & 0 \\  
 0 & 0 &-1 & 0 &+1 & 0 \\   
 0 & 0 & 0 &+1 & 0 &-1 \\  
 0 & 0 & 0 &-1 & 0 &+1 \\   
 \end{pmatrix}$}}
            $
            }
\caption{Default weight matrix $D$ for 3 input token vectors of dimension \textit{m}=2. 
Each row corresponds to the incoming weights of a hidden neuron and each column the outgoing weights of an input neuron. 
If there are more hidden neurons than pairs of input neurons times token vector dimension ($p \cdot m$), the additional rows contain only zeros.}
\label{fig:rbp4_2}
\end{figure}
%
%
As stated, Matrix $D$ is used for the loss function and also for initialisation. 
It is sufficient to use $D$ only in the loss function, but using $D$ also for initialisation speeds up convergence, as we found in initial tests, and is therefore applied in most of our experiments. 
In the initialisation, we first initialise weight matrix $W$ between the input and hidden layer as normal, e.g. with low-energy noise,  
and then copy the non-zero values of $D$ to $W$. 

%


 
Further, we define an ERBP loss based on the difference between $D$ and the actual weights in $W$. 
We compute the $L2$ or $L1$ norm of the difference $D-W$.
This loss term corresponds to a Bayesian prior on the weights with the mean defined by the values of $D$.
The $L2$ loss corresponds to a Gaussian prior and $L1$ loss to a Laplacian prior, such that back propagation maximises the posterior likelihood of the weights given the data \citep{williams1995bayesian}. 
The overall training loss $l_t$ is defined as 
\begin{equation}
    l_t = l_c  + \lambda \, l_{ERBP}
\end{equation}
where $l_c$ is the classification loss and $l_{ERBP}$ is 
\begin{equation}
    l_{ERBP} = wl_1 
     \sum_{i=1}^k (w_i - d_i)^2 + wl_2 
    \sum_{i=1}^k |(w_i - d_i)|.
\end{equation}
$\lambda$ is the regularisation parameter, corresponding to the inverse of the variance of the prior, effectively regulating the strength of the ERBP regularisation.
$(wl_1,wl_2)$ are the weights for L1 and L2, respectively. 
In principle, any combination of weights is possible, but here we only use $(1,0)$ or $(0,1)$ for $(wl_1,wl_2)$.
We call these methods ERBP L1 and ERBP L2 respectively.

For $l_c$ we have used cross entropy loss in the experiments, although it can be replaced with any other loss function suitable for the task.
The ERBP approach can be applied to feed froward and to recurrent networks. 
The only additional condition for recurrent networks is that we need to use a sliding window of size $n$.
ERBP requires $n(n-1)m$ neurons in the first hidden layer which leads to  $n(n-1)m^2$ weight parameters. 
If this number of neurons is present in the first hidden layer, no additional parameters are needed in the network.

Although the $L1$ loss 
encourages sparsity, that is not the specific goal of ERBP L1.   
Rather it is to enable learning of pairwise token equality by encouraging suitable features that compare token vectors in each dimension. 
Since equality learning is needed for abstract pattern recognition, but equality learning is a hard problem for neural networks, we hope that abstract pattern learning will benefit from ERBP similarly to RBP. 
We further apply this prior on a wide number of synthetic tasks which benefit from the equality based prior in different settings and experiments in the following sections.


\section{Experiments: Abstract Pattern Recognition}\label{sec:abstract_pattern}

In this section, we evaluate the ERBP method on generalising abstract patterns, based on equality relations, from training data. 
We compare ERBP to standard neural networks and Early and Mid-Fusion RBP, and we evaluate the effect on mixed abstract and concrete patterns. 

\subsection{Dataset}\label{subsec:data_gen}
For performing the rule learning experiments, we artificially generate data in the form of triples for each of the experiments. 
We generate triples in all five abstract patterns of length 3: AAA, AAB, ABA, ABB, and ABC for the experiments. 
We use as sample 
vocabulary $a ... l$ (12 letters) for both prediction and classification tasks. 
The sequences are divided differently for the different cases of classification. 
For all the experiments we used separate train, validation, and test sets with 50\%, 25\%, and 25\% of the data, respectively. 
All sampling (train/test/validation split, downsampling) is done per simulation. 
We used half of the vocabulary for training and the other half for testing and validation (randomly sampled). 
Using disjoint parts of the vocabulary means that the validation and test sets contain out of distribution patterns, in particular  non-zero activations of input neurons that are always zero in the training set. 

\subsection{Setup}\label{subsec:setup}
For all abstract pattern learning tasks, we used a standard feed-forward neural network. 
We used a grid search over hyperparameters: 
the number of epochs with values $[10,20,30]$,
and the number of neurons in the hidden layer with values $[10,20,30]$, except for ERBP where we use the minimal number necessary for the ERBP matrix.
We varied the regularisation parameter $\lambda$ with values $[0.01, 0.03, 0.1, 0.3, 1, 3, 10, 30]$ for the ERBP approaches.
For larger values of regularisation beyond 30, the performance did not improve and showed a downward trend.
We ran a total of 10 runs for each experiment. 
We used the Adam optimiser \citep{adam_ref} by default and also Stochastic Gradient Descent for comparison. 
We used a single hidden layer and a mini-batch size of 1 unless indicated otherwise. 
The networks have been implemented using the PyTorch  library\footnote{\url{http://pytorch.org}}. 



\subsection{Abstract Patterns}\label{subsec:abstract-patterns}
In this experiment, we aim to learn classification by abstract patterns based on equality as in the experiments by \cite{GaryMarcus1999}, using the terminology and tasks from \cite{weyde_kopparti_2019}. 
Triples of items $(\alpha,\beta,\gamma)$ following the abstract patterns AAA, AAB, ABA ABB and ABC are presented to the network. 
These abstract patterns can be described in logic using a binary equality predicate $eq(\cdot,\cdot)$.
E.g., the abstract patterns $ABA$ and $ABB$ can be represented by the following predicates:
\begin{align}
ABA&: \neg eq(\alpha,\beta) \land eq(\alpha,\gamma)
\\
ABB&: \neg eq(\alpha,\beta) \land eq(\beta,\gamma).
\end{align}
These predicates only depend on the equality relations and not the values of $\alpha,\beta, \text{and} \;  \gamma$. 
They are also called algebraic patterns or abstract rules \citep{GaryMarcus2001,dehaene2015neural}. 
We represent the input tokens to the network as one-hot encoded vectors and perform 5 different experiments as follows. 

\begin{enumerate}
    \item ABA vs other: In this task, one class contains only patterns of ABA and the other class contains all the other possible patterns (AAA, AAB, ABB, ABC), downsampled  for  class  balance. 
    Expressed in logic, the task is to detect whether $eq(\alpha,\gamma) \land \neg eq(\alpha,\beta)$ is true or false. 
    \item ABB vs other: In this task, one class contains only patterns of the form ABB and the other class contains all the other possible patterns (AAA, AAB, ABA, ABC), downsampled for class balance. 
    The logical task is to detect $eq(\beta,\gamma) \land \neg eq(\alpha,\beta)$.
    \item ABA-BAB vs other: In this task, class one contains patterns of type ABA as in 1. 
    For ABA sequences in the training set, the corresponding BAB sequences appear in the test set. 
    The other class contains all other possible patterns (AAA, AAB, ABB, ABC) downsampled per pattern for class balance as before. 
    \item ABA vs ABB: This task is like task 1 above, but only pattern ABB occurs in the other class, so that this task has less variance in the second class. 
    We expected this task to be easier to learn because two equality predicates $eq(\alpha,\gamma), eq(\beta,\gamma)$ change their values between the classes and are each sufficient to indicate the class. 
    \item ABC vs other: In this case, class one (ABC) has no pair of equal tokens, while the $other$ class has at least one of $eq(\alpha,\beta), eq(\alpha,\gamma), eq(\beta,\gamma)$ as $true$, i.e. detecting equalities without localising them is sufficient for correct classification. 
\end{enumerate}

\begin{table}[tb!]
\centering
\resizebox{\columnwidth}{!}{%
\begin{tabular}{lccccc}  
\hline
Type & 1) & 2) & 3) & 4) & 5) \\
\hline
Standard & 50 (1.86) & 50 (1.83) & 50 (1.73) & 50 (1.81) & 50 (1.68) \\
Early Fusion & 65 (1.26) & 65 (1.29) & 75 (1.22) & 55 (1.18) & 65 (1.04) \\
Mid Fusion & 100 (0.00) & 100 (0.00) & 100 (0.05) & 100 (0.00) & 100 (0.00) \\ 
ERBP L1 & 100 (0.00) & 100 (0.00) & 100 (0.02) & 100 (0.00) & 100 (0.00) \\
ERBP L2 & 100 (0.00) & 100 (0.00) & 100 (0.00) & 100 (0.00) & 100 (0.00) \\
\hline
\end{tabular}
}
\caption{ Abstract Pattern Learning. Test accuracy (in \%) with different models: (1) ABA vs other, 2) ABB vs other, 3) ABA-BAB vs other, 4) ABA vs ABB, 5) ABC vs other).  The table shows average and standard deviation (in brackets) over 10 simulations.}
\label{tab:exprs2}
\end{table}

In Table~\ref{tab:exprs2} (abstract patterns) we present test set accuracy, the training set accuracy was 100\% in all cases. 
We find that abstract patterns are almost perfectly generalised with ERBP L1 and ERBP L2, like with Mid Fusion RBP. 
We compared the differences between network performances across 10 simulations using the Wilcoxon Signed Ranked Test. 
The differences between ERBP L1/L2 and Standard as well as between ERBP L1/L2 and Early Fusion are statistically significant in all abstract pattern experiments at a threshold of $p < 0.05$. 
The differences between Mid Fusion, ERBP L1 and ERBP L2 are not significant.
However, while the Mid Fusion RBP achieves rounded 100\% test accuracy, the non-zero standard deviation indicates that some results are not fully accurate, while ERBP L2 has zero standard deviation in all tasks and ERBP L1 in all but one tasks. 

\subsection{Parameter Variations}\label{subsec:parvar}
We study the effect of several parameters: number of hidden layers, choice of optimiser, regularisation factor and  weight initialisation on the abstract pattern learning tasks. 

\begin{table}[tb!]
\centering
\begin{tabular}{cccccc}  
\hline
No of Hidden & Standard   & Early   & Mid  & ERBP L1 & ERBP L2  \\ Layers & Network  & Fusion  & Fusion  & & \\
\hline

h=2 & 50 (1.56) & 65 (1.23) & 100 (0.05) & 100 (0.00) & 100 (0.00) \\
h=3 & 52 (1.59) & 66 (1.08) & 100 (0.03) & 100 (0.00) & 100 (0.00)\\ 
h=4 & 55 (1.63) & 68 (1.12) & 100 (0.02) & 100 (0.00) & 100 (0.00) \\ 
h=5 & 56 (1.55) & 70 (0.89) & 100 (0.02)  & 100 (0.00) & 100 (0.00)\\
\hline
\end{tabular}
\caption{Network Depth. Test accuracy (in \%) for abstract pattern learning (ABA vs other) with \textit{h}= 2,3,4,5 hidden layers. Averages and standard deviations over 10 simmulations.}
\label{tab:hiddd}
\end{table}

\begin{table}[tb!]
\centering
\begin{tabular}{lrr}  
\hline
Type &  ERBP L1 & ERBP L2 \\
\hline
Adam   & 100 (0.00)  & 100 (0.00) \\
SGD     & 98 (0.06) & 96 (0.04) \\
\hline
\end{tabular}
\caption{Optimiser Choice. Test accuracy (in \%) and standard deviation of Abstract Pattern Learning (ABA vs other) using Adam and SGD for ERBP L1 and L2. 
Averages and standard deviations (in brackets) over 10 simulations. 
SGD can also lead to 100\% accuracy on the test set, but needs higher $\lambda$ values.}
\label{tab:opt14}
\end{table}

\textit{Network Depth}:
We test identity learning with deeper neural network models, using $h = {2,3,4,5}$ hidden layers. 
The results are listed in Table~\ref{tab:hiddd}, showing only minor improvements in the network performance for deeper networks. 
However, ERBP L1 and L2 generalisation is consistent and independent of the network depth.

\textit{Optimiser Choice}:
We test Stochastic Gradient Descent (SGD) and compare to the Adam optimiser \citep{adam_ref} for training the ERBP. 
We observe faster convergence and greater improvement in the overall accuracy with Adam compared to SGD.
We observe similar results for both ERBP L1 and L2. 
Table \ref{tab:opt14} summarises the results of abstract pattern learning for both optimisers with the regularisation parameter $\lambda$ set to 1.
We observe that the SGD does not reach full generalisation in this setting, however it does so at higher values of $\lambda$.

\begin{figure}[tb]
 \centerline{\includegraphics[width=9cm]
 {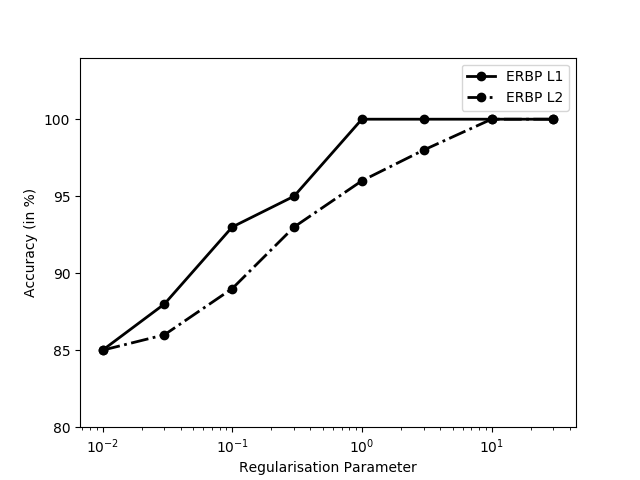}}
 \caption{Regularisation Factor. Test accuracy  of the network with ERBP L1 and L2 when varying the regularisation factor $\lambda$ (in logarithmic scale) for abstract pattern learning (ABA vs other).
 }
 \label{fig:example3_2}
\end{figure}

\textit{Regularisation Factor}:
We vary the regularisation factor $\lambda$ in the loss function of the ERBP model and observe that a large factor reliably leads to perfect generalisation in abstract pattern learning task. 
We vary the regularisation parameter $\lambda$ with values [0.01,0.03,0.1,0.3,1,3,10,30]. 
Figure \ref{fig:example3_2} shows how the effect depends on the size the regularisation factor $\lambda$ using L1 and L2 loss functions for abstract pattern learning task.
We test for larger values of $\lambda$, but the performance is worse beyond $\lambda$ value 30, hence, we do not consider values beyond that.

\textit{Weight Initialisation}:
We evaluate the effect of different weight initialisation schemes without the ERBP weight prior, to test whether ERBP  initialisation or other schemes might be sufficient on their own to learn abstract patterns. 
We test random, zero, Xavier \citep{xavier_init} and ERBP initialisation. 
For random initialisation, weight and bias are initialised from a uniform distribution over$(-1/\sqrt{n}, 1/\sqrt{n})$
where 
$n$ is the size of the weight matrix. 
The results of the experiment 
are shown in Table \ref{tab:init}.
Although ERBP initialisation shows the best generalisation in this comparison, it falls far short of the performance with the ERBP weight loss, which consistently achieves 100\% test accuracy (see Table~\ref{tab:exprs2}).

\begin{table}[tb]
\centering
\begin{tabular}{lrrrr}  
\hline
Initialisation: &  Random  & Zero  & Xavier  & ERBP (init only)  \\
\hline
Task 1 & 53 (1.82) & 51 (1.67) & 54 (1.53) & \textbf{61} (1.12)   \\
Task 2 & 53 (1.76) & 52 (1.72) & 54 (1.57) & \textbf{61} (1.09)   \\
Task 3 & 55 (1.73) & 54 (1.59) & 56 (1.49) &  \textbf{65} (1.04)  \\
Task 4 & 52 (1.89) & 54 (1.62) & 55 (1.62) & \textbf{60} (1.23)  \\
Task 5 & 52 (1.54) & 53 (1.58) & 54 (1.67) & \textbf{62} (1.20)  \\
\hline
\end{tabular}
\caption{Weight Initialisation. Test accuracy (in \%) with different weight initialisation methods on standard neural networks for various tasks ie. Task 1-5 Abstract Pattern Learning (ABA vs other, ABB vs other, ABA-BAB  vs other, ABA vs ABB, ABC vs other). 
Averages and standard deviations (in brackets) over 10 simulations.}
\label{tab:init}
\end{table}

We also evaluate the performance with the ERBP loss without the ERBP initialisation, using standard random weight initialisation instead. 
The ERBP loss on its own is sufficient for the model to produce 100\% accuracy when tested on Task 1 (ABA vs other). 
However, the training takes longer than with ERBP initialisation, doubling the number of epochs roughly from 30 to~60.

\subsection{Combined Abstract and Concrete Patterns}\label{subsec:mixed}


An important question about any inductive bias for abstract patterns is whether it has negative effects on other tasks. 
Here we test the effect of ERBP on concrete patterns. 
We will answer this question for real-world tasks in the following section.

Abstract pattern rules are independent of the actual values of $\alpha,\beta, \text{and} \;  \gamma$. 
Concrete patterns, on the other hand, are defined in terms of values from a vocabulary 
a,b,c, ... . 
E.g., sequences \textit{a**}, i.e. beginning with `a', or \textit{*bc}, ending with `bc', can be formulated in logic as follows:
\begin{align}
\textit{a**} &: eq(\alpha, \text{`a'}) \\
\textit{*bc} &: eq(\beta, \text{`b'}) \land eq(\gamma,\text{`c'}).
\end{align}

We conduct an experiment where the classes were defined by combinations of abstract and concrete patterns. 
Specifically we define four combined patterns based on combinations of the  abstract pattern  $ABA$ and $ABB$ with the concrete patterns  $a**$ and $b**$.  
E.g., the pattern $(ABA,a**)$ can be expressed logically as 
\begin{equation}
(ABA,\textit{a**}) =  eq(\alpha,\gamma) \land \neg eq(\alpha,\beta) \land eq(\alpha, \text{`a'}). 
\end{equation} 
We define four classes based on two abstract and two concrete patterns, where
each class is a combination of one abstract and one concrete pattern.
The classes are defined as follows:
\begin{align}
1: (\textit{ABA,a**}) \qquad 2&: (\textit{ABB,a**}) \\
3: (\textit{ABA,b**}) \qquad 4&: (\textit{ABB,b**}). 
\end{align}
We create the dataset with each class equally frequent and with items not prescribed by the patterns drawn at random, so that the abstract patterns (1,3 vs 2,4) and concrete  patterns (1,2 vs 3,4) are statistically independent. 
We use a vocabulary of 18 characters, out of which 12 are used for training and 6 are used for validation/testing in addition to `a' and `b', which need to appear in all sets because of the definition of the concrete patterns.
In our experiment we perform a four-way classification using  train, validation and test split of 50\%, 25\% and 25\% respectively. 
The results of the experiment are listed in Table~\ref{tab:mixed}

\begin{table}[tb]
\centering
\begin{tabular}{lr}  
\hline
Type &  Accuracy \\
\hline
Standard Network   & 39 (1.53)   \\
Early Fusion     & 66 (1.22)  \\
Mid Fusion    &  100 (0.00)  \\
ERBP L1   &   100 (0.00)   \\
ERBP L2 & 100 (0.00) \\
\hline
\end{tabular}
\caption{Combined Patterns. Test accuracy (in \%) for learning combined abstract and concrete patterns using different models.  
Averages and standard deviations (in brackets) over 10 simulations.
Both Mid Fusion and ERBP approaches result in 100\% accuracy in all cases.}
\label{tab:mixed}
\end{table}

The results show that the mixed patterns are perfectly generalised by Mid Fusion RBP and ERBP.
We also test with only concrete patterns  with two classes starting with $a**$ and $b**$ respectively, and observe that ERBP leads to 100\% accuracy unlike a standard network without ERBP which gives an accuracy of 94\%.
This shows that the (E)RBP does not impede neural network learning of concrete patterns. 
We try a range of values for $\lambda$ and found that a minimum was necessary for learning but higher values did not impede the learning of the concrete patterns. 

The models with ERBP lead to 100\% accuracy on a wide range of synthetic classification tasks.
Several authors, like \cite{seidenberg1999infants,vilcu2005two,shultz2006neural, frank2014getting,Alhama,Alhama2019} claimed success and showed improvements for the abstract pattern learning task.
However, the evaluation has mostly been conducted by testing whether the output of the network shows a statistically significant difference between the output for  inputs that conform to a trained abstract pattern and those that do not. 
From a machine learning perspective, this criterion is not satisfactory. 
Instead, we would expect that an identity/equality rule should always be applied correctly once if it has been learned from examples, at least in cases of noise-free synthetic data, which we have achieved in with the ERBP loss.  
\section{Experiments: Language and Music Modelling}
\label{sec:realworld}
In order to evaluate the effect of ERBP on language models, we perform experiments with real world data on character and word prediction in natural language and also on pitches in melodies. 
We use RNN, GRU and LSTM models for character, word, and musical pitch prediction. 

The input 
vectors represent items in a  context window of length \textit{n}. 
Therefore there are $n \cdot (n-1) / 2 $ pairs of vectors to compare. 
The ERBP is applied in the same way as before in the first hidden layer after the input. 
In these tasks, the encoding of the information is not binary, as in the synthetic tasks, but integer-based or continuous. 
Thus, the differences in DR units and the ERBP hidden neurons represent not just equality but a form of distance of the input values, which we expect to be useful for prediction.

\subsection{Character and Word Prediction}
We use Wikitext-2 \citep{DBLP:journals/corr/MerityXBS16}
for character and word prediction. 
For word prediction, the entire dataset with 2 million words was used with  train/validation/test split as in the original dataset with pretrained Glove embeddings \citep{pennington2014glove} with 50 dimensional vectors.
For character prediction, a truncated version with $60\,000$ words was used with a train/valida\-tion/\linebreak[0]{}test split of 50/25/25 and with integer encoding of the characters.

\begin{table}[tb!]
\centering
\begin{tabular}{lrrrr}  
\hline
Type & n=10  & n=20 & n=30 & n=40   \\ 
\hline
RNN &
 21.99 & 21.73 & 21.42 & 21.47 \\
Early Fusion & 
 21.90 &  21.64 &  21.35 & 21.24 \\ 
Mid Fusion & 
 21.89  &  21.60 & 21.33   & 21.20  \\
ERBP L1 & 
 20.66 & \textbf{20.50} & 20.04 & 20.32 \\
ERBP L2 & 
 \textbf{20.60} & 20.53 & \textbf{20.03} & \textbf{20.27} \\
\hline
\end{tabular}
\caption{Character prediction. Perplexity per character 
for various context lengths \textit{n} = 10,20,30,40 with RNNs on the Wikitext-2 dataset truncated to 60k words with integer encoding tested on standard RNN, Early, Mid Fusion, ERBP L1 and L2.
The best results (in bold) occur for the ERBP approaches with $\lambda = 0.3$.
}
\label{tab:char1}
\end{table}

We use networks with 2 hidden layers with 50 neurons each.
Both the hidden layers have recurrent connections and we have used this network structure throughout for all the experiments.
We train the networks for up to 30 epochs for characters and up to 40 for words. 
The learning rate was set to 0.01. 
Training typically converges after less than 30 epochs and we select the best model according to the validation loss.
We test with regularisation parameter $\lambda$ values [0.01,0.03,0.1,0.3,1,3] for ERBP. 
We test only for the above set of  $\lambda$ values, as the performance when $\lambda > 3$ was worse for both character and word predictions tasks.
Hence, we did not consider larger values of $\lambda$.
We use a context length $n$ of [10,20,30,40] and evaluate again ERBP L1 and L2, RBP Early and Mid Fusion, and standard networks.

\begin{table}[tb!]
\centering
\begin{tabular}{lrrrr}  
\hline
Type & 
n=10 & n=20 & n=30 & n=40   \\ 
\hline
GRU  & 
21.97 & 21.72 & 21.39  & 21.45 \\
Early Fusion & 
21.89 & 21.63 & 21.33 & 21.20 \\
Mid Fusion   & 
21.87 & 21.59 & 21.30 & 21.19 \\
ERBP L1 & 
20.61 & \textbf{20.43} & 20.02 & 20.25\\
ERBP L2 & 
\textbf{20.53} & 20.46 & \textbf{19.96} & \textbf{20.23} \\
\hline
\end{tabular}
\caption{Character prediction. Perplexity per character for various context lengths \textit{n} = 10,20,30,40 with GRUs on the Wikitext-2 dataset truncated to 60k words with integer encoding tested on standard GRU, Early, Mid Fusion, ERBP L1 and L2.
The best results (in bold) occur for the ERBP approaches with $\lambda = 0.3$.
}
\label{tab:char3}
\end{table}

\begin{table}[htb!]
\centering
\begin{tabular}{lrrrr}  
\hline
Type & 
n=10 & n=20 & n=30 & n=40   \\ 
\hline
LSTM  & 
21.88 & 21.64 & 21.34  & 21.33 \\
Early Fusion & 
21.84 & 21.63 & 21.27 & 21.14 \\
Mid Fusion   & 
21.81 & 21.57 & 21.24 & 21.16 \\
ERBP L1 & 
20.54 & \textbf{20.32} & 19.89 & \textbf{20.18}\\
ERBP L2 & 
\textbf{20.45} & 20.33 & \textbf{19.87} & 20.19 \\
\hline
\end{tabular}
\caption{Character prediction. Perplexity per character for various context lengths \textit{n} = 10,20,30,40 with LSTMs on the Wikitext-2 dataset truncated to 60k words with integer encoding tested on standard LSTM, Early, Mid Fusion, ERBP L1 and L2.
The best results (in bold) occur for the ERBP approaches with $\lambda = 0.3$.
}
\label{tab:char2}
\end{table}

The results for character prediction and word prediction are summarised in Tables \ref{tab:char1}, \ref{tab:char3}, \ref{tab:char2} and \ref{tab:word1}, \ref{tab:word3} \ref{tab:word2} for RNN, GRU and LSTM models respectively. 
The performance was best for LSTMs with ERBP and the choice of $\lambda$ value didn't have a major effect in the case of word prediction, unlike character prediction where $\lambda = 0.3$ gave best performance. 
We test for larger values of the regularisation factor but the performance did not improve.
\begin{table}
\centering
\begin{tabular}{lrrrr}  
\hline

Type & n=10 & n=20 & n=30 & n=40   \\ 
\hline

RNN  & 
186.97 & 164.56  &  114.26 & 101.95\\
Early Fusion & 
186.80 & 164.52 & 114.21 &  101.89 \\
Mid Fusion   &
186.52  & 164.41 & 114.19 & 101.69 \\
ERBP L1 & 
185.83 & 163.98 & \textbf{113.96} & 100.25 \\
ERBP L2 & 
\textbf{185.54} & \textbf{163.90} & 113.99 & \textbf{100.23} \\
\hline
\end{tabular}
\caption{Word Prediction. Perplexity per word for various context lengths \textit{n} = 10,20,30,40 with RNNs on the Wikitext-2 dataset consisting of 2 million words with integer encoding tested on standard RNN, Early, Mid Fusion, ERBP L1 and L2.
The best results (in bold) occur for the ERBP approaches.
}
\label{tab:word1}
\end{table}

\begin{table}
\centering
\begin{tabular}{lrrrr}  
\hline

Type & n=10 & n=20 & n=30 & n=40   \\ 
\hline
GRU  &
174.62  & 160.31 & 114.05 & 97.62\\
Early Fusion & 
174.35  & 160.26 & 113.86 & 96.39\\
Mid Fusion   & 
174.42  & 160.27 & 113.55 & 96.35\\
ERBP L1 & 
173.33  & 159.36 & 113.42 & \textbf{96.21} \\
ERBP L2 & 
\textbf{173.21}  & \textbf{159.32} & \textbf{113.32} & 96.32 \\
\hline
\end{tabular}
\caption{Word Prediction. Perplexity per word for various context lengths \textit{n} = 10,20,30,40 with GRUs on the Wikitext-2 dataset consisting of 2 million words with integer encoding tested on standard GRU, Early, Mid Fusion, ERBP L1 and L2. The best results (in bold) occur for the ERBP approaches.
}
\label{tab:word3}
\end{table}

\begin{table}
\centering
\begin{tabular}{lrrrr}  
\hline

Type & n=10 & n=20 & n=30 & n=40   \\ 

\hline
LSTM  &
160.44  & 153.61 & 113.40 & 93.28\\
Early Fusion & 
160.32  & 153.42 & 113.26 & 93.23\\
Mid Fusion   & 
160.38  & 153.30 & 113.23 & 93.15\\
ERBP L1 & 
159.29  & \textbf{152.78} & 113.02 & 92.96 \\
ERBP L2 & 
\textbf{159.22}  & 152.89 & \textbf{112.99} & \textbf{92.82} \\
\hline
\end{tabular}
\caption{Word Prediction. Perplexity per word for various context lengths \textit{n} = 10,20,30,40 with LSTMs on the Wikitext-2 dataset consisting of 2 million words with integer encoding tested on standard LSTM, Early, Mid Fusion, ERBP L1 and L2.The best results (in bold) occur for the ERBP approaches.
}
\label{tab:word2}
\end{table}

We use the Wilcoxon Signed Rank test with threshold $p=0.05$ as above to compare the standard models with ERBP L1 and L2 for both character and word prediction across 60000 words for context length $n=10$. 
For character prediction, RNN, GRU and LSTM with ERBP L1 and L2 are significantly better than the standard RNN, GRU and LSTM models. 
For word prediction, all ERBP models except LSTM ERBP L1 are significantly better than the corresponding standard models. 

In the case of language modelling for both character and word level, our results are not comparable to the SOTA models on the Wikitext-2 dataset, 
which have about 395M parameters in perplexity of 34.1 without pre-training \cite{wang2019language}. 
Even with state-of-the-art LSTM models our experiment are not comparable 
as we used a subset of Wikitext-2. 

\subsection{Melody Prediction}
In another experiment, we test ERBP for predicting the pitch of the next note in melodies with a selection of the Essen Folk Song Collection \citep{Schaffrath1995}, which comprises of 8 sets of songs in different styles. 
Pitches are integer-encoded. 
With RNN, GRU and LSTM we use two hidden layers of size 20. 

\begin{figure}[tb]
 \centerline{\includegraphics[width=5.49cm]
 {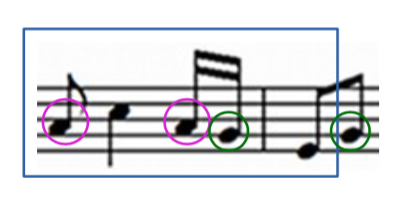}}
 \caption{Examples from the Essen Folk Song Collection : Repetition of notes in a given context length `n`' = 6 here represented in a blue rectangle. Pink circles correspond to repeating notes in the context length `n'=6. Green circles correspond to repetition of notes within a given context and the next note being predicted.}
 \label{fig:rep2}
\end{figure}

In Figure \ref{fig:rep2} we indicated the sliding window at the initial position as a blue rectangle which is the given context.
We also highlighted repeated notes, where we hypothesise that the repetition has a bearing on the prediction of the next note, if the network learns to detect the repetition. 
The notes in the pink circle correspond to the repetitions within a given context, and the notes in the green circle are an example of repetition between a given context and a corresponding note.

We use Adam and tested values [0.01, 0.03, 0.1, 0.3, 1, 3] for the regularisation factor $\lambda$.
Similar to the character and word prediction experiments, we evaluate for smaller $\lambda$ values, as the performance degraded for values of $\lambda > 3$.
We tested context lengths of 5 and 10 and the train/validation/test split was set to 50/25/25\%.

\begin{table}[htb!]

\centering
\begin{tabular}{lrrr}  
\hline
Type        & RNN    
& GRU 
& LSTM   \\ 
\hline
Standard  & 2.8012 
& 2.7968 
& 2.7928 \\
Early Fusion & 2.7365 
& 2.7328 
& 2.7314 \\ 
Mid Fusion & 2.7356 
& 2.7302 
& 2.7219 \\ 
ERBP L1 & 2.7264 
& \textbf{2.7252} 
&  \textbf{2.7132}  \\
ERBP L2 & \textbf{2.7240} 
& 2.7260  
& 2.7144   \\
\hline
\end{tabular}

\caption{Pitch Prediction. Average cross entropy per note for pitch prediction using context length $n=5$ on the Essen Folk Song dataset using standard models, Early, Mid fusion and ERBP approaches.
The best results (in bold) are achieved with the ERBP approaches at $\lambda$ = 0.3.
}
\label{tab:mel1}
\end{table}

\begin{table}[htb!]
\centering
\begin{tabular}{lrrr}  
\hline
Type & RNN    
& GRU   
& LSTM   \\ 
\hline
Standard  & 2.7502  
& 2.7457 
& 2.7432 \\
Early Fusion & 2.6735 
& 2.6723 
& 2.6712 \\ 
Mid Fusion & 2.6653  
& 2.6650 
&  2.6648\\ 
ERBP L1 & \textbf{2.6642} 
& \textbf{2.6641} 
&  2.6639  \\
ERBP L2 & 2.6649
& 2.6645  
& \textbf{2.6632}    \\
\hline
\end{tabular}

\caption{Pitch Prediction. Average cross entropy per note for pitch prediction using context length  $n=10$ on the Essen Folk Song dataset using standard models, Early, Mid fusion and ERBP approaches.
The best results (in bold) are achieved with the ERBP approach at $\lambda$ = 0.3.
}
\label{tab:mel2}

\end{table}

Tables~\ref{tab:mel1} and \ref{tab:mel2} summarise the results of the neural networks without and with RBP in early, mid fusion and using ERBP L1 and L2. 
We observe the best results for $\lambda$ values of 0.3. 
Larger values of $\lambda$ did not result in improved results.
We have tested for statistical significance over the results for the 8 subsets, using the Wilcoxon signed rank test with threshold $p = 0.05$ as above. 
We find that all model types (Standard/Early/Mid/ERBP L1/L2) are different with statistical significance, so that ERBP L1 and L2 outperform standard networks and RBP models for this task.

In the case of music modelling, SOTA model on monophonic pitch prediction task is the work by \citep{DBLP:journals/corr/abs-1709-08842} in which a feature discovery \textit{PULSE} learning framework has been proposed.
The cross entropy was $2.395$ and the best performance with ERBP using LSTMs in our experiments was $2.663$.
However, the models cannot be directly compared here as well, as the approach, number of features, model architecture, parameters, context length are not directly comparable.
ERBP model has also not been evaluated on longer context lengths, which we believe might result in improved or closer SOTA model performance when fine-tuned effectively.

\section{Experiments: Complex Learning Tasks}\label{exp:complex}

In this section, we evaluate the use of ERBP for more complex tasks. 
Specifically, we experiment with graph distance learning as an example of distance and relation learning and with compositional sentence entailment as a compositional natural language task. 

\subsection{ERBP for learning Graph Edit Distance}
In this experiment, we extend the scope from equality based patterns and sequential prediction to a situation of end-to-end learning of a distance function. 
Our goal is to learn a distance function between graphs. 
Given two graphs $G_1, G_2$, a graph distance model is a function $f(G_1, G_2)$ that computes a scalar distance value. 
This experiment uses Graph Neural Networks (GNN), 
which are popularly used in relational learning tasks.
In a GNN, a graph structure with nodes and edges is used as input. 
A standard GNN model comprises of two encoders to learn node and edge representations using iterative message passing. 
For more details about the GNN model refer \cite{DBLP:journals/corr/abs-1812-08434}.

Our model is based on embedding each graph into a vector with a GNN and then classify pairs of graphs based on their distance. 
We use the graph edit distance, which counts the number of operations needed for transforming one graph into another.
For more details refer to \citep{DBLP:journals/corr/abs-1904-12787}.
A graph distance model that aligns with the GED is trained by giving graphs with small edit distance a low distance and otherwise high distance as learning targets.

For our dataset, random pairs of graphs are generated from the Graph Similarity data generator \citep{DBLP:journals/corr/abs-1904-12787}. 
The generator yields pairs of graphs $(G_1, G_2)$ and labels $t\in\{-1, 1\}$ for this pair where `-1' and and `+1' signify that GED is above or below a threshold, respectively. 
The loss function we optimise is  
\begin{equation}
L_\mathrm{pair} = \mathbb{E}_{(G_1, G_2, t)}[\max\{0, \gamma - t(1 - d(GNN(G_1), GNN(G_2)))\}] .
\end{equation}
This loss encourages similar graphs to have distance smaller than $1-\gamma$, and dissimilar graphs to have distance greater than $1 + \gamma$, where $\gamma$ is a margin parameter, the NN is applied twice with shared weights, and $d$ is a distance function to compare the graph.
The dimensionality of node vectors is 32 and that of graph embedding vectors is 128. 
We evaluate the model here with several distance functions \textit{d}: The standard Euclidean distance and trainable NNs, specifically a Siamese Network and an MLP with and without ERBP .
The MLP has a single hidden layer with 256 neurons. 
The whole network is trained end to end with a total of 1000 graph pairs in a split of 60/20/20\% for training, validation and test set, respectively. 
The results are shown in Table \ref{tab:opt11}.

\begin{table}[tb!]
\centering
\begin{tabular}{lrr}  
\hline
Distance Function &  Accuracy \\
\hline
Standard MLP & 95.61\% \\
Siamese network   & 96.09\% \\
Euclidean distance &  97.54\%  \\
MLP with ERBP  &   97.72\%  \\
\hline
\end{tabular}
\caption{Graph Edit Distance. Accuracy of GNNs with different distance functions for the graph edit distance (GED) task.}
\label{tab:opt11}
\end{table}

As can be seen from Table \ref{tab:opt11}, the GNN outperforms a Siamese network in the GED task.
GNN with MLP+ERBP results in further improvements when compared to a standard GNN, even though the difference is small.
Table \ref{tab:opt11} shows that standard GNN reaches accuracy of 97.54\% and GNN with ERBP produces 97.72\% on 1000 pairs generated from the Graph Similarity data generator \citep{DBLP:journals/corr/abs-1904-12787}.
There is no directly comparable state of the art for this experiment and the standard Euclidean distance produces already good results. However, the MLP with ERBP reduces the number errors by over 9\% compared to Euclidean distance (2.28 vs 2.46\%), while the other two trainable models perform worse than Euclidean.


\subsection{Compositional Sentence Entailment}
In another experiment, the effect of ERBP is tested on an entailment task in natural language data.
In this task, we extend the notion of abstract rules to the semantics of sentences, albeit in a very restricted syntactic context.
The notion of compositionality here is in the sense of understanding of how words combine to form a sentence in a way that generalises to words and sentences that have not previously been encountered.


The dataset used in the task is the Compositional Comparison Dataset introduced by \cite{comp}.
The task is to classify a pair of sentences into neutral, contradiction or entailment originally motivated from the popular Stanford Natural Language Inference (SNLI) dataset \cite{snli:emnlp2015} natural language inference classification task.
In this abstract compositionality task, pairs of sentences (A,B) are generated which differ by permutation of words, such that each of the pairs represent different relations.
This dataset is called as the Compositional Comparison dataset as all the pairs of sentences are compositionally generated by changing the word ordering in the sentences.
There are three categories `Same',`More-less' and `Not'. 
The structure of the sentences used is shown in Table \ref{tab:abs_sen}, where the type assignment holds true for any X, Y and Z. 
It is tested if the neural network models to generalise to X, Y and Z that have never been encountered before. 

\begin{table}[tb!]
 \begin{center}
 \begin{tabular}{lrrr}
  \hline
  Type & Entailment & Contradiction & No of Pairs  \\
  \hline
  Same &  X is more Y than Z & Z is more Y than X & 14670 \\
  More-Less  & Z is less Y than X & X is less Y than Z & 14670 \\
    \hline
 \end{tabular}
\end{center}
 \caption{Sentence Entailment. Comparisons dataset summary. The premise to compare to is: X is more Y than Z}
 \label{tab:abs_sen}
\end{table}

In this experiment, 
we excluded the 'not' category and the 'neutral' class from the original dataset,
to focus on challenging tasks that are suitable for testing compositionality based on comparisons.
The values for X, Y and Z can be arbitrary nouns for X and Z and adjectives for Y. 

For example, in the \textit{Same Type} Contradiction case, the sentence pairs differ only in the order of the words. \\
\textit{A} : The woman is more cheerful than the man \\
\textit{B} : The man is more cheerful than the woman \\
And in the \textit{Entailment} case, the sentences are identical: \\
\textit{A} : The woman is more cheerful than the man \\
\textit{B} : The woman is more cheerful than the man
(Entailment) 

In the \textit{More-Less Type}, the \textit{Contradiction} pairs differ by whether they contain the word `more' or the word `less': \\
\textit{A} : The woman is more cheerful than the man \\
\textit{B} : The woman is less cheerful than the man \\
There the \textit{Entailment} also differs additionally by word order: \\
\textit{A} : The woman is more cheerful than the man \\
\textit{B} : The man is less cheerful than the woman 

We encode the sentences pairwise by word in A and B, i.e. one pair of words at same position in A and B per time step. 
We apply the ERBP prior to the pair of words rather than using a time window as in Section 5. 
We use Glove embeddings \citep{pennington2014glove} and a network with a single LSTM layer and a sigmoid layer for classification as the output. 
The dataset consists of approximately 5k sentences with equal split between the categories. 
As a baseline, we use a Bag of Words (BOW) encoding that averages the Glove embeddings with a Multi-Layer Perception (MLP).
The state of art model for this type of task is the InferSent model \cite{DBLP:journals/corr/ConneauKSBB17}.

\begin{table}[tb!]
\centering
\begin{tabular}{lrr}  
\hline
Model &  Same type & More-Less type \\
\hline
BOW-MLP   & 50\% & 30.24\% \\
InferSent model \cite{DBLP:journals/corr/ConneauKSBB17}  & 50.37\% & 50.35\%    \\
LSTM with ERBP  & 51.32\% & 50.64\%    \\
\hline
\end{tabular}
\caption{Sentence Entailment. Accuracy of various models for abstract compositionality task.}
\label{tab:opt1}
\end{table}

The results are listed in Table \ref{tab:opt1}.
The BOW/MLP model results are exactly at chance level and the InferSent model perform only slightly above at 50.37\% for the `Same' type class.
LSTM with ERBP results improves by just under 1pp with 51.32\%.
For the `More-Less' type, the Bag of Words MLP model gives only about 30\% accuracy where as the InferSent and the LSTM model with ERBP give slightly above 50\% accuracy.

We can say that the model did benefit from the ERBP and the improvements were consistent for the both the `Same' and `More-Less' types but the improvement size is small.

\section{Discussion}
\label{discussion}
We have introduced ERBP as a re-modelling of RBP in the form of a Bayesian prior on the weights as a solution for abstract pattern learning that is simple to integrate into standard recurrent and feed-forward network structures as a regularisation term and that retains the full flexibility of the network learning without hard coded weights. 

The experiments in Section~\ref{sec:abstract_pattern} show that the ERBP L1 and L2 models are effective in learning classification of abstract patterns on artificial sequence data, and they improve performance slightly over the original RBP.
The learning of abstract patterns with ERBP is robust over a wide range of parameter settings (Section \ref{subsec:parvar})
and does not limit the network's learning of concrete pattern in combination with abstract patterns or on its own (Section~4.5).

The second point of this study is whether the enabling of the abstract pattern learning can improve neural network learning on real-world data. 
This was addressed in the second set of experiments on language and music modelling as well as a sentence entailment task and graph comparison.
The language and music modelling results show that the prediction performance benefits from ERBP in all tested tasks of character and word prediction in natural language, and pitch prediction in melodies. 
We observe that ERBP with recurrent models (RNN, GRU, LSTM) leads to consistent improvements across different context lengths and in tasks when compared to RBP and standard RNN model types.
The gain in perplexity obtained with ERBP is small but in most cases statistically significant compared to RBP and standard NNs.

We performed two additional experiments on more complex tasks, compositional sentence entailment with LSTMs and graph comparison with graph neural networks, where we achieved small improvements with ERBP. 
For the compositional entailment task, and LSTM network with ERBP outperforms the baseline Bag of Words model and the state-of-the-art InferSent encoder-decoder model \citep{DBLP:journals/corr/ConneauKSBB17} slightly.
In the Graphs Edit Distance task, the GNN with MLP/ERBP as a trainable distance function outperforms both a Siamese network and a GNN with Euclidean distance, but the difference between the two GNN variants is small. 
While we believe that modelling of equality and distance is a necessary component to better modelling, these tasks seem to require more comprehensive approaches where the comparisons are more strategically applied. 
There is clearly room for improvement, particularly in the entailment task. 

Overall, our results show that ERBP as a weight prior is an easy method to integrate into the standard networks, wherever there are multiple input vectors to compare. 
We demonstrated the use of ERBP with several neural networks models, including feed-forward, RNNs and their gated variants, and Graph Neural Networks. 
All networks did benefit from ERBP, with dramatic improvements in abstract pattern recognition on synthetic sequence data. 
The prediction improvements are relatively smaller real-life applications, pointing to the fact that learning equality is not the only aspect that is missing for improved language learning. 
This was even more pronounced in the complex tasks of entailment and graph distance. 
Overall, we see the results provides evidence that addressing systematicity in the form of equality and distance learning in neural networks is not just a theoretical issue, but that it has a positive impact on real-life tasks. 

From a practical perspective, ERBP is straightforward to implement and integrate into all standard network architectures, so that is still worth exploring, even if the gains are not always substantial. 
We have observed no negative effects of ERBP compared to standard networks and the method proved to be robust to parameter values and the choice of L1 and L2 in all the experiments.

\section{Conclusions}
\label{conclusions}
In this study we focused on abstract pattern learning with neural network architectures based on equality and distance relations.  
We introduce Embedded Relation Based Patterns (ERBP), a prior on the weights of feed-forward and recurrent neural network architectures that leads to learning of abstract patterns, where standard neural networks fail. 
ERBP has been proven in our experiment as a way of creating an inductive bias that encourages learning of abstract patterns in synthetic as well as real word data. 
The ERBP weight prior acts as a regularisation term which is easy to integrate into standard back-propagation.


We observed in experiments that with ERBP L1 and L2 weight priors, neural network models learn generalisable solutions to classification tasks based on abstract pattern learning. 
While most approaches to learning abstractions focus on the structure of the network and non-standard operators, this is to our knowledge the first approach that addresses abstract pattern recognition with a prior on the network weights. 


In general, we see ERBP-like mechanisms as a way of modelling inductive bias in neural network architectures that helps improve data efficiency by systematic learning. 
We believe that this approach of creating an inductive bias with weight priors can be extended easily to other forms of relations and will be beneficial for many other learning tasks.

The effect of ERBP will be studied further to assess its effectiveness on different tasks and datasets. 
Beyond its current form, ERBP can be adapted to other forms of abstract relations 
and improving generalisation on more complex tasks such as question answering or perception-based reasoning by extending ERBP and combining it with memory models.
We also see promise in extending this approach towards learning higher-level abstractions by combining ERBP with explicit memory and thus approaching a more comprehensive learning model.

\bibliographystyle{spbasic}
\bibliography{references}

\end{document}